\pdfoutput=1

\documentclass[11pt]{article}

\usepackage[]{acl}

\usepackage{times}
\usepackage{latexsym}
\usepackage{graphics}
\usepackage{graphicx}
\usepackage{subcaption}
\usepackage{soul}

\usepackage[T1]{fontenc}

\usepackage[utf8]{inputenc}

\usepackage{microtype}

\usepackage{booktabs}

\title{Analyzing Gender Representation in Multilingual Models}

\author{Hila Gonen\textsuperscript{1} \, Shauli Ravfogel\textsuperscript{2,3} \, Yoav Goldberg\textsuperscript{2,3}\\
\textsuperscript{1}Paul G. Allen School of Computer Science \& Engineering, University of Washington \\
\textsuperscript{2}Computer Science Department, Bar Ilan University \\
\textsuperscript{3}Allen Institute for Artificial Intelligence \\
  {\tt  \{hilagnn, shauli.ravfogel, yoav.goldberg\}@gmail.com}
  }

\usepackage{fancyhdr}
\fancypagestyle{firststyle}
{
    \fancyhf{}
    \fancyhead{}
    \lhead{Published at the Workshop on Representation Learning for NLP (RepL4NLP 2022)}
}

\begin{document}
\maketitle

\thispagestyle{firststyle}

\begin{abstract}

Multilingual language models were shown to allow for nontrivial transfer across scripts and languages. In this work, we study the structure of the internal representations that enable this transfer. We focus on  the representation of gender distinctions as a practical case study, and examine the extent to which the gender concept is encoded in \textit{shared subspaces} across different languages. Our analysis shows that gender representations consist of several prominent components that are shared across languages, alongside language-specific components. The existence of language-independent and language-specific components provides an explanation for an intriguing empirical observation we make: while gender classification transfers well across languages, interventions for gender removal, trained on a single language, do not transfer easily to others. 

\end{abstract}

\section{Introduction}

Pretrained models of contextualized representations \cite{elmo,bert,roberta} are known in their ability to capture both explicit and implicit information during training. A special case of these models are multilingual models \cite{bert,xlmr}, which are pretrained with texts in multiple languages. These models were shown to induce the emergence of similar representations in different languages, a phenomenon that was put to use for transfer between languages in end-tasks \cite{PSG19,MSS20,mbert}. However, the underlying mechanism is still not clear, and we do not know yet the full extent to which the representations of these models share information across languages.

The rise of pretrained models has been accompanied with growing concern regarding sensitive information they might encode, e.g. gender or ethnic distinctions. Pretrained language models were shown to be sensitive to gender information, both when it is explicitly stated in texts, as well as when it can be inferred from implicit information \cite{gender-ctx,gender-encoders}. We still lack a complete understanding of what the model captures, and the ways to control and change the information in this context as well.

In this work, we aim to shed light on the way gender, a popular use case of a human-interpretable concept, is represented in multilingual models, and whether it is encoded in a language-dependant way. In a series of experiments, we uncover a surprising finding: gender-identification ability is highly transferable across languages (section \ref{sec:probes}) but neutralizing gender identification is not (section \ref{sec:debias}). While these two findings may seem contradictory at first glance, this is explained by several levels of gender marking: both cross-lingual and language-specific (section \ref{sec:analyze}). 

We start our analysis by training gender classifiers and examining their ability to transfer across languages. We then proceed to identifying ``gender subspaces'' --- subspaces that encode gender --- in each language, with the goal of understanding which information is language-specific, and which is shared across languages. Following recent work on linear interventions \cite{inlp,amnesic,rc-counterfactuals, ravfogel2022linear}, we take an ``amnesic'' approach: we study the extent to which \textbf{neutralizing} the gender subspace in one language interferes with gender prediction in another language. Finally, we analyze the similarity in the gender-encoding components across languages. 

We find that while linear probes for gender transfer well between languages --- that is, a gender classifier that is trained on one language predicts gender well in another language, the method we employ for neutralizing gender fails to transfer across languages. A deeper analysis reveals a fine-grained organization of the gender-encoding subspaces across languages: they are spanned by a few main directions, which are largely similar across languages; but in addition to these directions, there are other directions that are language-specific.
The existence of several similar directions explains the high degree of transferability of linear gender classifiers across languages, while the existence of a large amount of language-specific information explains the inability to efficiently remove gender information in one language based on another language's representation.  

We summarize our findings and contributions as follows: (a) we show that gender-identification is highly transferable across languages (Section~\ref{sec:probes}); (b) we find that neutralizing gender identification does not transfer well across languages (Section~\ref{sec:debias}); (c) we demonstrate that gender subspaces are spanned by a few directions that are largely similar across languages; and also by other directions that are language-specific (Section~\ref{sec:shared}); (d) we find that the directions that are shared across languages are the most dominant ones (Section~\ref{sec:dominant}).

The code for our experiments is available at \url{https://github.com/gonenhila/multilingual_gender}.

\section{Related Work}

\paragraph{Multilingual Representation Analysis}

\newcite{PSG19} begin a line of work that studies mBERT's representations and capabilities. They inspect the model's zero-shot transfer abilities using different probing experiments, and propose a way to map sentence representations in different languages, with some success. \newcite{KWM20} further analyze the properties that affect zero shot transfer of bilingual BERTs. \newcite{WD19} perform transfer learning from English to 38 languages, on 5 different downstream tasks and report good results. \newcite{WCG19} learn alignment between contextualized representations, and use it for zero shot transfer. \newcite{phillip_mbert} make an attempt to control different aspects of mBERT and identify those that contribute the most to its transfer ability. 

Beyond focusing on zero-shot transfer abilities, an additional line of work studies the representations of mBERT and the information it stores. Using hierarchical clustering based on the CCA similarity scores between languages, \newcite{SMS19} are able to construct a tree structure that faithfully describes relations between languages. \citet{mbert-syntax} learn a linear syntax-subspace in mBERT, and point out to syntactic regularities in the representations that transfer across languages.
In \newcite{CKK19}, the authors define the notion of \textit{contextual} word alignment and show improvement in zero-shot transfer after fine-tuning accordingly. In \newcite{LRF20}, the authors assume that mBERT's representations have a language-neutral component, and a language-specific component and provide an experimental setting to partially support this assumption. Finally, in \newcite{mbert}, the authors propose an explicit \emph{decomposition} of the representations to language-encoding and language-neutral components, and also demonstrate that implicit word-level translations can be easily distilled from the model when exposed to the proper stimuli. 

Unlike previous works, we pay attention specifically to how gender is manifested in the representations, as a case study for the analysis of a concrete societal property. We do that by focusing on the information included in the representations themselves, rather than on downstream tasks.

\paragraph{Gender Representation in Multilingual Models}

To the best of our knowledge, no previous work focuses on the way gender is represented in multilingual models and the extent to which such representations are shared across languages, with the single exception of \newcite{AB22}, who showed that information about gender---as well as other morphological attributes---is partially encoded in the same neurons across languages.

Some work has been done on identifying and mitigating gender bias in languages other than English \cite{ZSZ19,BNG20}. \newcite{grammatical} identify and debias a new type of gender bias, unique to gender-marking languages. \newcite{gg_ryan} look at the relationships between the grammatical genders of inanimate nouns and their co-occurring adjectives and verbs.  In \newcite{spanishbias}, the authors suggest a method for converting between masculine-inflected and feminine-inflected sentences in morphologically rich languages, and use them for counterfactual data augmentation in order to reduce gender stereotyping. \newcite{mbios} analyze gender bias in multilingual word embeddings, and evaluate it intrinsically and extrinsically. They point to several factors that influence the gender bias in multilingual embeddings, among which are the pretrained monolingual word embeddings, and the alignment method used. Additionally, \newcite{philipp_debias} focus on contextualized embeddings, analyze the gender representation in BERT, and also put efforts into English-Chinese cross lingual debiasing. Finally, \newcite{multi_ind} focus on Indian languages when debiasing multilingual embeddings.

\section{Datasets and Multilingual Representations}
\label{Sec:data}

For our experiments we use the BiosBias Dataset \cite{biosbias}, the Multilingual BiosBias Dataset \cite{mbios} and the multilingual BERT model (mBERT, \cite{bert}) as detailed below.

\paragraph{Multilingual Gender Data.}

\newcite{biosbias} collected the English BiosBias
dataset, a set of short-biographies written in third person, and annotated by perceived gender. To do so they identified online biographies, written in English, from Common Crawl, by filtering for lines that match a pattern of a name and an occupation.\footnote{A sequence of two capitalized words followed by “is a(n) (xxx) \textit{title},” where \textit{title} is a profession from BLS Standard Occupation Classification system.} Gender is labeled using heuristics, based on names and pronouns. In their work, they have demonstrated that profession classifiers trained on this dataset condition on the gender concept, resulting in fairness issues. \newcite{mbios} evaluate the bias in cross-lingual transfer settings, for which they have created the Multilingual BiosBias (MLBs) Dataset which contains a similar set of biographies in three additional languages: French, Spanish and German. Note that these are not translations of the English portion, but are crawled independently with a similar method. 

For our experiments we use both datasets, so that we have biographies in English, Spanish and French.\footnote{Since the datasets are not available online, we used the scripts the authors provide for crawling them ourselves. The German portion we were able to extract was too small, so we decided to avoid experimenting with it.} To decrease noise, we filter out examples of professions with less than 500 occurrences. Table \ref{tab:stats} describes the statistics of the dataset in all languages. Note that the dataset is not balanced with respect to gender, especially for French and Spanish (same as before our filtering), and that the English portion is significantly larger. Following \cite{biosbias}, we split randomly into Train/Dev/Test sets with ratio of 65\%/10\%/25\%, while ensuring that the main class (professions) is balanced across them. Unfortunately, biographies data for more languages is not available at this point, so we opt to use English, French and Spanish only.

\begin{table}[h!]
    \centering
    \resizebox{\columnwidth}{!}{
    \begin{tabular}{cccccc}
    \toprule
     &  examples &  female &   male & majority &  \# prof \\
    \midrule
       En &    255682 &  118344 & 137338 & 53.71 &          28 \\
       Fr &     42773 &   12196 &  30577 & 71.49 &          19 \\
       Es &     46931 &   12867 &  34064 & 72.58 &          27 \\
    \bottomrule
    \end{tabular}}
    \caption{Statistics of the MLBs dataset.}
    \label{tab:stats}
\end{table}

\paragraph{Multilingual Representations.}
To study the representation of the gender concept in a multilingual setting, we use multilingual BERT (mBERT,\footnote{Implemented with HuggingFace \cite{hugging}.} 110M parameters) \cite{bert}. For each example in the dataset, we extract its representation from mBERT by averaging the last-layer representations in context of all the tokens in the paragraph.

\section{Gender Representation across Languages}
\label{sec:transfer}

\subsection{Transfer of Gender Probes}
\label{sec:probes}

As a first step in understanding gender representation in multilingual models, we start with a basic experiment that aims to evaluate the extent to which gender is represented similarly across languages. The goal of this experiment is to check whether features that help predict the gender of a contextualized representation in one language are also predictive of gender in another language.

To this end, we train a linear classifier (logistic Regression classifier, trained in SKlearn\footnote{https://scikit-learn.org/stable/} with default parameters) for gender classification in a \textsc{source} language, and use it as is to predict the gender in a \textsc{target} language. The training is done over the mBERT representations of the training examples (see Section \ref{Sec:data}).

The results, presented in Table~\ref{tab:cls_transfer}, indicate that gender classifiers transfer very well across languages, with only a slight degradation in performance when applied in a different language. For example, the accuracy of the English gender classifier in-language is 99.27\%, and when the French or Spanish classifiers are used to predict gender in the English data, the accuracy is 98.10\% and 97.29\%, respectively. The same trend is observed for the French and Spanish datasets. These results suggest that gender information is linearly accessible in mBERT representations and is shared between languages.

\begin{table}[h!]
    \centering
    \scalebox{0.8}{
    \resizebox{\columnwidth}{!}{
    \begin{tabular}{c|ccc}
    \toprule
    & En train & Fr train & Es train \\ 
    \midrule
    En test & \textbf{99.27} & 98.10 & 97.29 \\ 
    Fr test & 95.97 & \textbf{97.50} & 94.61 \\ 
    Es test & 84.04 & 84.10 & \textbf{85.97} \\
    \bottomrule
    \end{tabular}}}
    \caption{Accuracy of gender classification across languages with linear classifiers. Rows represent the language of the prediction data, columns represent the language in which the classifier was trained.}
    \label{tab:cls_transfer}
\end{table}

\subsection{Cross-lingual Linear Gender Removal}
\label{sec:debias}

The experiment described above suggests that some gender components are shared between languages. As bias mitigation techniques focus on the \textit{removal} of gender information, a natural question that arises is whether mitigation efforts trained on one language would transfer to another. This question is important for two reasons. First, if possible, this has a potential practical utility -- e.g., enabling bias mitigation in low-resource languages, for which training data is scarce. Second, the degree of success in transfer of bias mitigation efforts is a complementary way to assess whether the representation of gender is indeed multilingual.

Previous experiments on removing the gender concept from neural representations show encouraging results in-language for English. These are done using INLP \cite{inlp}, an existing approach for the identification and neutralization of ``concept subspaces'', e.g. the gender concept. In these experiments, \newcite{inlp} show they manage to neutralize the ability of linear probes to recover gender information from the representations. In light of the above results that show high quality \textit{transfer} of gender classifiers \textbf{across} languages, we leverage the INLP method, and attempt to \textit{remove} gender information from the representations \textbf{across} languages.

Note that the goal of the following experiment is not \textit{debiasing} gender but rather \textit{analyzing} gender directions across langauges -- INLP is used in this experiment as an analysis tool, rather than a debiasing tool. In what follows, we give an overview of INLP, and then describe the experiment and its results.

\paragraph{Iterative Nullspace Projection (INLP)}

INLP \citep{inlp} aims to remove linearly-decodable information from vector representations. 

INLP constructs a concept subspace iteratively, by finding directions of the relevant concept (e.g. gender) and neutralizing them by projecting the representations onto their nullspace. On each iteration, a classifier is trained on the representations, which were projected onto the nullspace of the previous classifiers, i.e., the classifier is optimized to identify \emph{residual} information which was not captured by previous directions. This iterative procedure relies on the intuition that in order to find a subspace whose neutralization \emph{hinders} the ability to predict some concept, one first needs to identify the directions that \emph{encode} that concept, and only then neutralize them.

Formally, given a dataset of representations $X$  (in our case, mBERT representations) and annotations $Z$ for the information to be removed (gender) the method renders $Z$ linearly unpredictable from $X$. It does so by iteratively training linear predictors $w_1, \dots, w_n$ of $Z$, calculating the projection matrix onto their nullspace $P_{N} := P_N(w_1), \dots,  P_N(w_n)$, and transforming $X \gets P_{N}X$. By the nullspace definition, this guarantees $w_iP_{N}X=0, \forall w_i$, i.e., the features that $w_i$ uses for gender prediction are neutralized. Note that the guarantee is only with respect to linear separation.

While the nullspace $N(w_1, \dots, w_n)$ is a subspace in which $Z$ is not linearly predictable, the complement rowspace $R(w_1, \dots, w_n)$ is a subspace of the representation space $X$ that corresponds to the property $Z$. In our case, the nullspace is the \emph{gender neutral subspace} and the rowspace is the \emph{gender subspace}. As part of the analysis in this work, we utilize INLP in two complementary ways: (1) we use the \textit{nullspace} projection matrix $P_N$ to zero out the gender subspace, in order to render the representations gender-neutral,\footnote{to the extent that gender is indeed encoded in a linear subspace, and that INLP finds this subspace.} this projection is onto the \textbf{gender-neutral subspace}; and (2) we use the \textit{rowspace} projection matrix $P_R = I - P_N$ to project mBERT representations onto the \textbf{gender subspace}, keeping only the parts that are useful for gender prediction.

\paragraph{Method}

We start by training INLP in one language (En, Fr or Es) and identifying the complementing subspaces: the gender-neutral subspace -- \textit{nullspace}, and the gender subspace -- \textit{rowspace} (the latter is used in Section \ref{sec:analyze}). We then neutralize the gender subspace in \textit{another} language. Finally, we examine the influence of this intervention and asses the effect of gender information reduction. Importantly, the directions are \textbf{learned} by INLP and are not predefined according to a word list or in any other manual manner.

We run INLP with the objective of identifying the gender, with SVM classifiers (using SKlearn) for 100 iterations.\footnote{as we have noticed that 100 iterations are enough to remove gender information in-language for all three languages.} We use the average representations of the training paragraphs (averaging over the final-layer in-context representations of all tokens).

\paragraph{Results}

Tables~\ref{tab:gender} and ~\ref{tab:prof} depict the results of gender and profession predictions (with Logistic Regression) in each language (rows) before and after applying INLP (each column stands for a different language for training INLP). In-language, the accuracy of gender prediction drops to majority after applying INLP, while profession classification is only slightly hurt. For example, for English we get gender prediction accuracy of 53.7 compared to 99.3 before applying INLP, and profession prediction accuracy of 78.1 compared to 79.9 before applying INLP. Note that this is the \textbf{expected} behaviour as a result of applying INLP, since INLP is designed to remove as much information as possible for the \textit{guarded attribute}, namely gender, with minimal effect on the main task. Indeed \cite{inlp} show the same result for English in the original paper. However, across languages, there is virtually no effect, both for gender prediction and profession prediction. For example, English gender and profession predictions drop from 99.3 to 98.1 and from 79.9 to 79.5, respectively, after applying Spanish INLP. This result is surprising in light of the high quality transfer of gender identification across languages shown in the previous experiment (Section \ref{sec:probes}, Table~\ref{tab:cls_transfer}).

Interestingly, the largest drops in performance of profession classification due to application of INLP are in-language. This can be explained by the inherent correlations between gender and profession signals -- removing gender information hurts the ability to predict the profession in the same language. This is not the case across languages since, as seen by the gender prediction results, gender information is not removed from the representations when applying INLP across languages.

\begin{table}[h!]
    \centering
    \begin{tabular}{l|r|rrr}
    \toprule
     &  before &  En INLP &  Fr INLP &  Es INLP \\
    \midrule
       En &   99.3 &    53.7 &    97.6 &    98.1 \\
       Fr &   97.8 &    95.1 &    71.4 &    94.9 \\
       Es &   85.7 &    82.8 &    82.6 &    72.5 \\
    \bottomrule
    \end{tabular}
    \caption{Gender prediction before and after applying INLP. Rows stand for the language in which we predict, columns stand for the language in which we train INLP. We use 100 iterations of INLP in each language.}
    \label{tab:gender}
\end{table}

\begin{table}[h!]
    \centering
    \begin{tabular}{l|r|rrr}
    \toprule
     &  before &  En INLP &  Fr INLP &  Es INLP \\
    \midrule
       En &   79.9 &    78.1 &    79.2 &    79.5 \\
       Fr &   73.0 &    72.4 &    68.2 &    72.4 \\
       Es &   57.8 &    57.1 &    57.3 &    51.8 \\
    \bottomrule
    \end{tabular}
    \caption{Profession prediction before and after applying INLP. Rows stand for the language in which we predict, columns stand for the language in which we train INLP. We use 100 iterations of INLP in each language.}
    \label{tab:prof}
\end{table}

\section{Analyzing the Cross-linguality of Gender Representation}
\label{sec:analyze}

At first glance, the two results presented in the previous section look contradicting: linear gender classification transfers well across languages while gender removal using INLP does not. In this section we provide a detailed analysis that accounts for this discrepancy and sheds light on the arrangement of gender in multilingual representations -- this is essentially the main result of this work. Under this more fine-grained view we present, we see that gender representation is neither shared between languages nor unique per language, but is actually only partially shared between languages. This allows for some transferability (as seen in Section~\ref{sec:probes}), but prevents gender removal across languages (as seen in Section~\ref{sec:debias}).

To define the term ``partial sharing'' formally, we represent gender in each language as a collection of linear directions that together span the gender subspace of that language. This collection of directions can be identified using INLP: when training INLP in a specific language, we get a sequence of orthogonal linear classifiers that are able to predict gender with a decreasing level of accuracy, with the first classifier being the most accurate one. Together, these directions define the gender subspace of the language. This formulation allows us to more easily analyze the extent to which gender is similarly encoded across languages. 

We hypothesize that the two aforementioned results are compatible because \textbf{some of these gender directions are shared between languages, while others are language-specific}. The shared directions allow high quality transfer of gender classification across languages, while the language-specific directions allow gender prediction even after applying INLP cross-lingually since they are not identified in the source language. 
In what follows, we devise two experiments to verify this hypothesis and quantify this phenomenon.

\subsection{Shared Gender Directions across Languages}
\label{sec:shared}

\paragraph{High Level Description and Intuition}

In the following experiment we analyze the relation between gender representations in the different languages. For that we leverage the formulation of gender representation as a collection of many different directions in the space. We aim to answer the following question: \textit{are gender directions fully shared across languages, fully disjoint, or split (i.e. some are shared across languages and some are disjoint)}?

Concretely, in order to derive a measure of overlap between two given subspaces, we measure the effect of neutralizing the gender subspace of one language, on the total variance in the gender subspace of \emph{another} languages; intuitively, the larger the overlap is between the gender subspaces in both languages, the larger the drop in variance is expected to be.

\paragraph{Method}

Given two languages $A$ and $B$ we propose the following pipeline: (i) project the representations of language $A$ \emph{onto its gender subspace} in order to discard information that is not predictive of gender in that language; (ii) project the already-projected representations \emph{onto the gender-neutral subspace} of language $B$ in order to remove the gender-information captured in the subspace of language $B$; (iii) measure the drop in the total variance of the representations of language $A$ between steps i and ii.

To draw a more fine-grained view of the transfer of gender-neutralization, in step iii we perform Principle Component Analysis (PCA), and record the total variance explained by the first $n$ principle components. Thus, we ask not only how does the gender-neutralizing in language $B$ affect the gender subspace of language $A$, but also \emph{which} PCA directions are affected. Concretely, we plot the total explained variance by the first $n$ principle components. If the intervention does not change the plot at all, this means that the two gender subspaces are completely orthogonal, and if the variance drops to zero at once, this means that the two gender subspaces are completely aligned.

\paragraph{Compared Representations}

We start by training INLP and obtaining a collection of 100\footnote{We use 100 for each language even when INLP required less iterations to converge, so as to be consistent across languages and avoid artifacts due to the number of dimensions.} gender directions in each language (En, Fr and Es), from the most prominent to the least prominent one.  We compare different sets of representations as detailed below, for English vs. French, English vs. Spanish and French vs. Spanish (the explanation below is assuming English vs. French):

\begin{itemize}
    \item \textsc{orig}: Original representations (in English).
    \item \textsc{EnGender}: \textsc{orig} projected on the English gender subspace (rowspace).
    \item \textsc{EnRand}: \textsc{orig} projected with a random matrix with the same dimensions as the EnGender matrix (for comparison).
    \item \textsc{EnGender+FrNeutral}: \textsc{EnGender} projected on the French gender-neutral subspace (nullspace).
    \item \textsc{EnGender+FrRand}: \textsc{EnGender} projected on a random matrix with the same dimensions as the French gender-neutral matrix (for comparison).
    \item \textsc{EnGender+EnNeutral}: \textsc{EnGender} projected on English gender-neutral subspace (nullspace, as a sanity check).
\end{itemize}

\paragraph{Result Analysis}

\begin{figure}[h!]
\centering
    \subfloat[English and French.]{
    \label{fig:pca_en_fr}
    \includegraphics[scale=0.48]{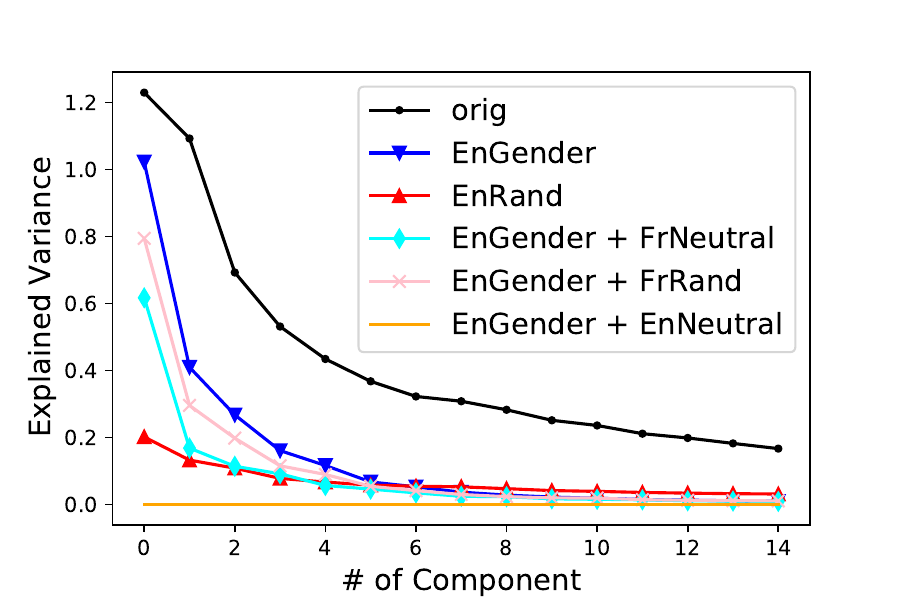}
    } \\
    \subfloat[English and Spanish.]{
    \label{fig:pca_en_es}
    \includegraphics[scale=0.48]{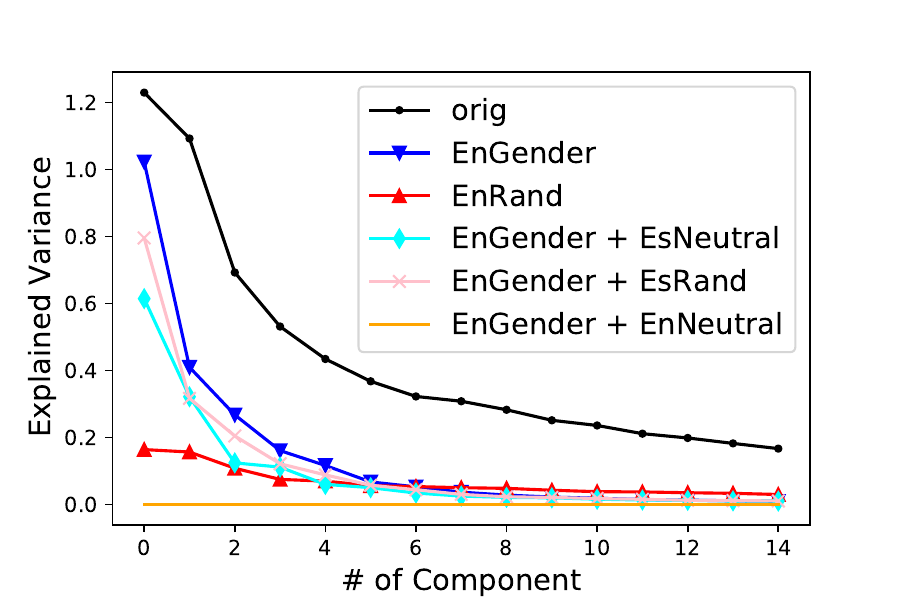}
    } \\
    \subfloat[French and Spanish.]{
    \label{fig:pca_fr_es}
    \includegraphics[scale=0.48]{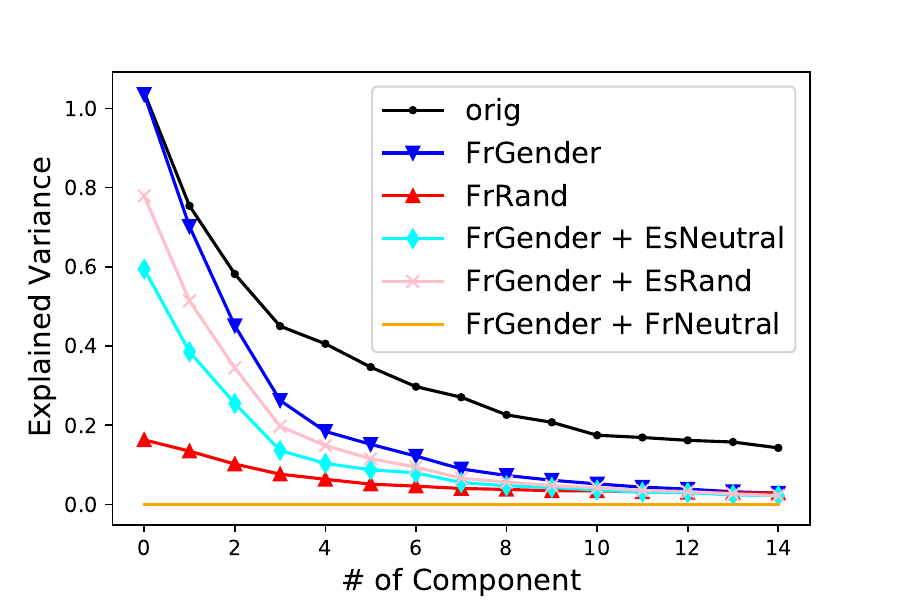}
    }
    \caption{Explained variance of PCA of different representations, for all three language pairs.}
    \label{fig:pca}
\end{figure}

The results are shown in Figure~\ref{fig:pca}. The plots support our initial hypothesis: indeed, we find that gender directions are shared between languages, but only partially. Focusing on English vs. French, we can see that as expected, the curve of \textsc{EnGender+FrNeutral} (cyan) is lower than that of \textsc{EnGender} (blue), implying that there are shared gender directions between English and French. Recall that projecting the representations on the English gender subspace (\textsc{EnGender}) keeps mainly English gender directions, and then projecting on the French gender-neutral subspace (\textsc{EnGender+FrNeutral}) removes French gender directions. If no directions are shared, this should result with similar values for both \textsc{EnGender} and \textsc{EnGender+FrNeutral}. However, the sharing is only partial: if all directions are shared, we expect \textsc{EnGender+FrNeutral} to be zero (similar to \textsc{EnGender+EnNeutral}), which is not the case.

\paragraph{Controls}

The \textsc{EnGender+FrRand} projections are intended as reference for \textsc{EnGender+FrNeutral}. If there are shared gender directions between English and French, we expect the curve of \textsc{EnGender+FrNeutral} to be lower than that of \textsc{EnGender+FrRand}, since by projecting on the French gender-neutral subspace we are expected to lose more information than with a random projection with the same dimensions. In Figure~\ref{fig:pca_en_fr} we see that the curve of \textsc{EnGender+FrNeutral} (cyan) is indeed lower than that of \textsc{EnGender+FrRand} (pink), indicating that the loss of information is not due to random shared directions.

Note also that the curve of \textsc{EnGender} (blue) is significantly higher than that of \textsc{EnRand} (red). We hypothesize that this is due to the fact that gender is usually dominant in natural texts, especially in a dataset that includes information about individuals, as this one. Thus, keeping only gender information by projecting on the English gender subspace keeps a large portion of the information, compared to projecting on arbitrary directions of the same dimension.

Another sanity check is obtained by projecting \textsc{EnGender} on the English gender-neutral subspace (\textsc{EnGender+EnNeutral}), this should, by definition, result in a 0 line, which is indeed the case (orange).

\subsection{Similarities of Dominant Directions}
\label{sec:dominant}

In the previous section we established the hypothesis that some gender directions are shared between languages while others are language-specific. Now, we turn to perform a more fine-grained analysis where we look at the specific directions in the different languages.

We look at the first 100 classifiers (trained during INLP) in two languages, and compute all pairwise cosine similarities between them (across languages). This leads us to a surprising result -- only the \textbf{first} classifiers in both languages are similar to each other, while the rest are not: we get that the 3 highest similarities are between the first English classifier and the first French classifier, between the second English classifier and the second French classifier, and between the third English classifier and the third French classifier, with values of 0.777, 0.597 and 0.453, respectively. For comparison, the average absolute cosine similarity among all pairwise similarities of the first 100 classifiers in English and French is 0.037. This result means that not only are some directions shared cross-lingually while others are not, but also that the most dominant directions are those that are shared, while the less predictive directions are those that are language specific.

\begin{figure}[h!]
\centering
    \subfloat[Similarity between the $i^{th}$ classifiers in En and Fr.]{
    \label{}
    \includegraphics[scale=0.48]{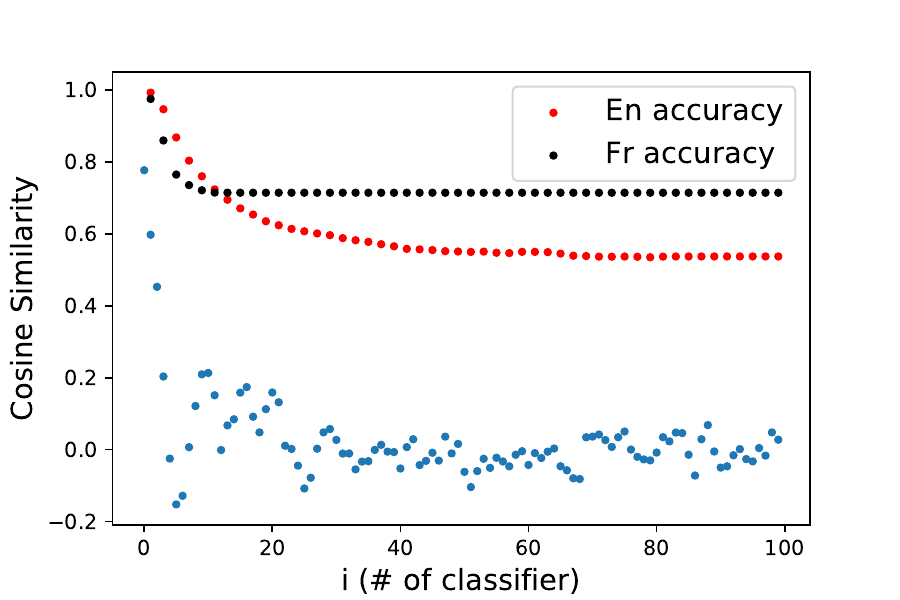}
    } \\
    \subfloat[Similarity between the $i^{th}$ classifiers in En and Es.]{
    \label{}
    \includegraphics[scale=0.48]{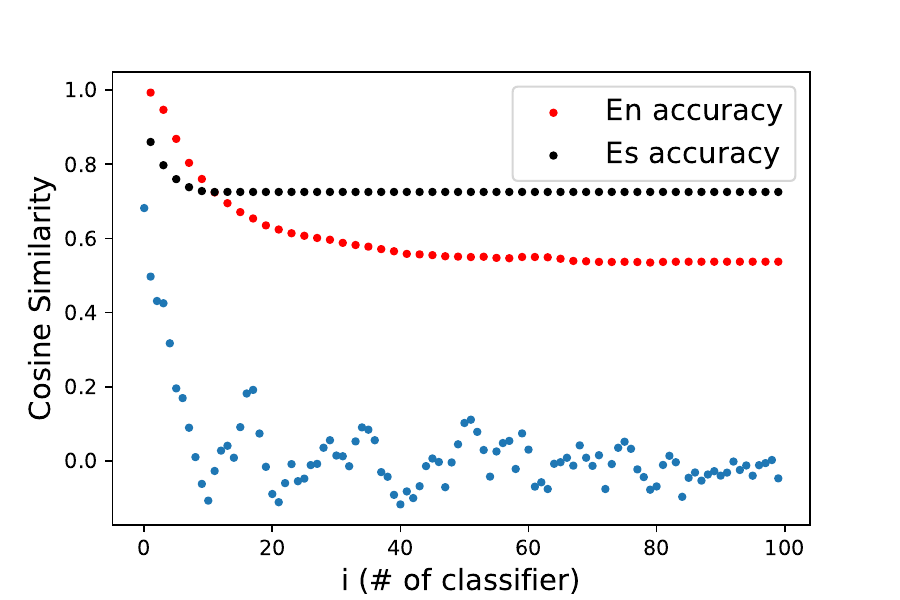}
    } \\
    \subfloat[Similarity between the $i^{th}$ classifiers in Fr and Es.]{
    \label{}
    \includegraphics[scale=0.48]{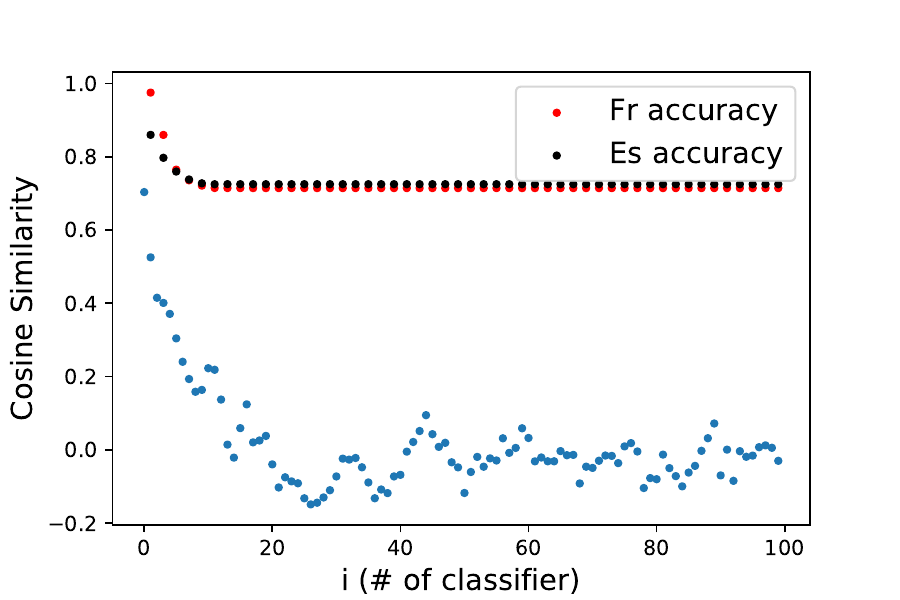}
    }
    \caption{Similarity between the $i^{th}$ classifiers (blue) in all three language pairs. The gender classification accuracy in-language (black and red) is added for reference.}
    \label{fig:sim_classifiers}
\end{figure}

Figure~\ref{fig:sim_classifiers} depicts the similarities of the $i$th classifiers for the two languages (English-French, English-Spanish and French-Spanish). We also plot the gender classification accuracy in-language for reference. This result completes the picture and serves as an explanation for the extremely high quality transfer of gender classification across languages -- the most dominant directions that represent gender in each languages are cross-lingual, which enables high accuracy in zero-shot transfer of linear gender classifiers across languages. However, less dominant gender directions are language specific, but are predictive enough so as to prevent gender neutralization across languages using INLP.

\subsection{Accuracy across Languages}

\begin{figure}[h!]
\centering
    \subfloat[Gender prediction accuracy in English.]{
    \label{fig:cross-lang-en}
    \includegraphics[scale=0.48]{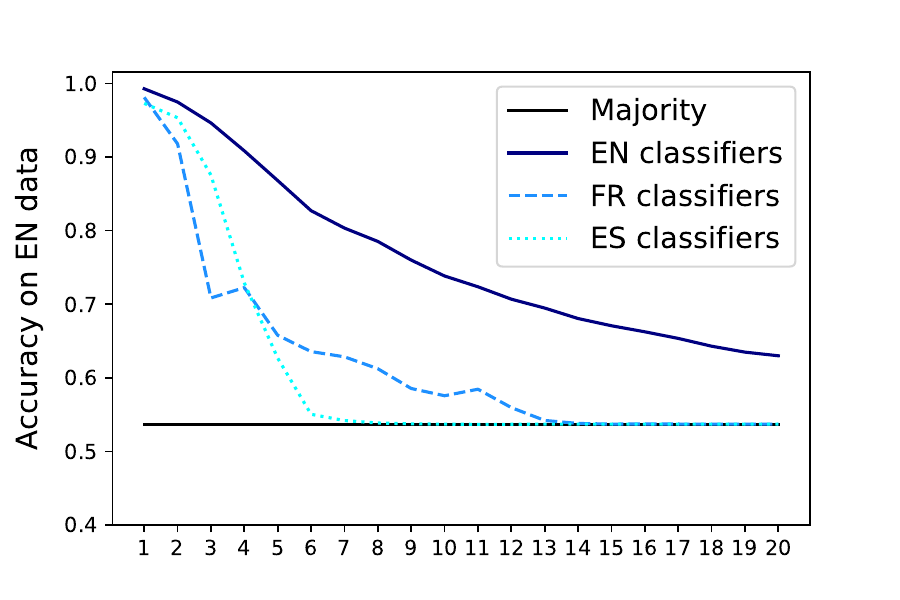}
    } \\
    \subfloat[Gender prediction accuracy in French.]{
    \label{fig:cross-lang-fr}
    \includegraphics[scale=0.48]{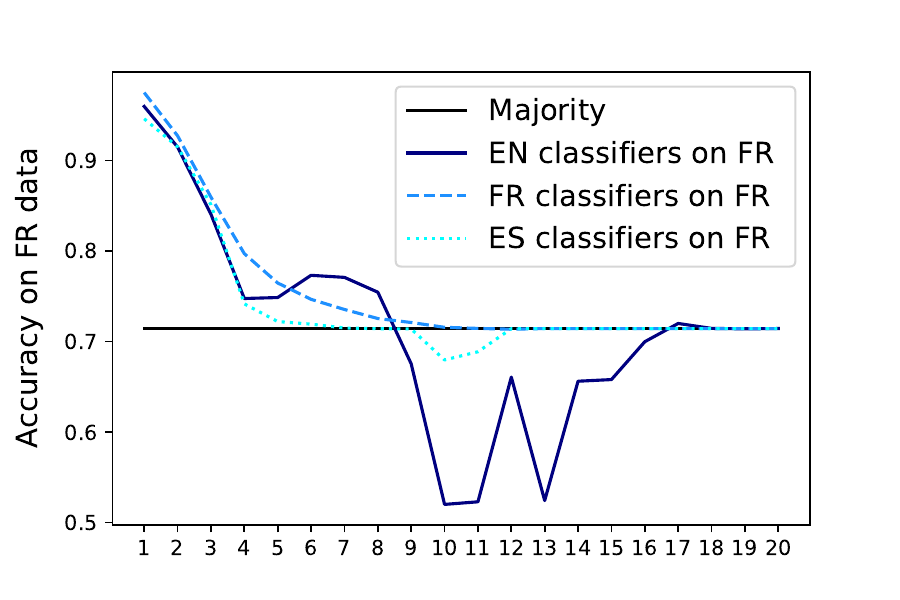}
    } \\
    \subfloat[Gender prediction accuracy in Spanish.]{
    \label{fig:cross-lang-es}
    \includegraphics[scale=0.48]{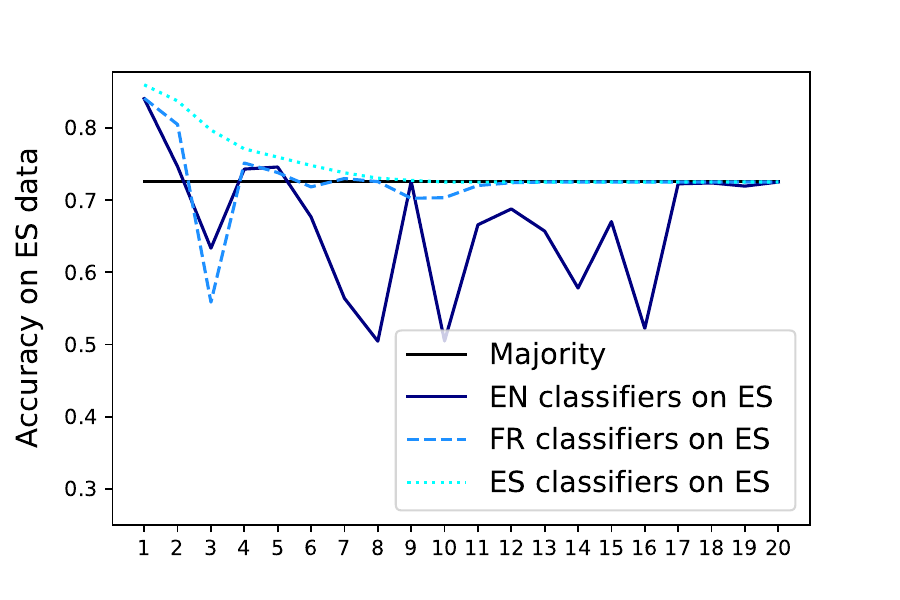}
    }
    \caption{Gender prediction accuracy with the different classifiers in- and across-languages.}
    \label{fig:cross-lang}
\end{figure}

Finally, we also look at the performance of each classifier (trained during INLP) across languages. In Figure~\ref{fig:cross-lang}, we depict the gender prediction accuracy in-language and across languages. We consistently get that the performance of the first 2-3 classifiers trained in-language and also across languages is relatively similar, with a significant divergence between in-language and across languages training for the subsequent classifiers. This matches the observation of high similarity only between the first classifiers across the different languages.

\section{Conclusion}

Towards better understanding of the underlying mechanism of multilingual modeling, in this work we focus on the way gender is represented across languages. We analyze and quantify the extent to which gender information is shared in multilingual representations in English, French and Spanish.

We find that on the one hand, gender prediction transfers very well across languages: training a linear classifier on English data yields a high quality classifier for French and Spanish as well (true for all three languages in both directions). On the other hand, our attempt to transfer gender removal in cross-lingual manner was unsuccessful. 

We show that these two results are compatible, and together they shed light on the structure of the representation space: we provide experimental evidence that the most salient directions are shared between languages (enabling good transfer of the classifiers), while others are unique per language (interfering with gender removal across languages). The key observation is that a \textit{single} ``good'' direction of the gender subspace in one language is enough for cross-lingual gender prediction transfer, while transfer of gender neutralization requires \textit{all} directions to be shared, otherwise, the remaining ones can be used to recover gender information after the removal of the shared ones. 

\section{Ethical Considerations}
\label{Sec:ethics}

Gender bias mitigation has attracted a lot of attention as a practical and socially important field of study. This paper contributes to this effort by studying the internal organization of gender representations. We note that gender and bias are complicated and multi-faceted constructs. When studying gender bias in neural models, we unavoidably rely on a narrow notion of binary gender, as reflected in several annotated datasets. As such, we see this study as a preliminary attempt that is based on a relatively narrow concept of gender, that does not reflect the subtle ways by which gender bias is manifested. We advise for caution when applying the conclusions of this study to other notions of gender or different definitions of bias. 

We acknowledge that gender is not a binary property. Due to lack of existing resources, we use binary gender as a rough approximation of reality. We hope to account for this in future work.

\section*{Acknowledgements}

We would like to thank Ran Levy for valuable ideas and feedback. This project received funding from the Europoean Research Council (ERC) under the Europoean Union's Horizon 2020 research and innovation programme, grant agreement No. 802774 (iEXTRACT).

\bibliography{all}

\begin{thebibliography}{32}
\expandafter\ifx\csname natexlab\endcsname\relax\def\natexlab#1{#1}\fi

\bibitem[{Antverg and Belinkov(2022)}]{AB22}
Omer Antverg and Yonatan Belinkov. 2022.
\newblock On the pitfalls of analyzing individual neurons in language models.
\newblock In \emph{ICLR}.

\bibitem[{Bansal et~al.(2021)Bansal, Garimella, Suhane, and
  Mukherjee}]{multi_ind}
Srijan Bansal, Vishal Garimella, Ayush Suhane, and Animesh Mukherjee. 2021.
\newblock Debiasing multilingual word embeddings: A case study of three indian
  languages.
\newblock In \emph{Proceedings of the 32nd ACM Conference on Hypertext and
  Social Media}, pages 27--34.

\bibitem[{Bartl et~al.(2020)Bartl, Nissim, and Gatt}]{BNG20}
Marion Bartl, Malvina Nissim, and Albert Gatt. 2020.
\newblock \href {https://aclanthology.org/2020.gebnlp-1.1} {Unmasking
  contextual stereotypes: Measuring and mitigating {BERT}{'}s gender bias}.
\newblock In \emph{Proceedings of the Second Workshop on Gender Bias in Natural
  Language Processing}, pages 1--16, Barcelona, Spain (Online). Association for
  Computational Linguistics.

\bibitem[{Cao et~al.(2019)Cao, Kitaev, and Klein}]{CKK19}
Steven Cao, Nikita Kitaev, and Dan Klein. 2019.
\newblock Multilingual alignment of contextual word representations.
\newblock In \emph{International Conference on Learning Representations}.

\bibitem[{Chi et~al.(2020)Chi, Hewitt, and Manning}]{mbert-syntax}
Ethan~A. Chi, John Hewitt, and Christopher~D. Manning. 2020.
\newblock \href {http://arxiv.org/abs/2005.04511} {Finding universal
  grammatical relations in multilingual {BERT}}.
\newblock \emph{CoRR}, abs/2005.04511.

\bibitem[{Conneau et~al.(2020)Conneau, Khandelwal, Goyal, Chaudhary, Wenzek,
  Guzm{\'a}n, Grave, Ott, Zettlemoyer, and Stoyanov}]{xlmr}
Alexis Conneau, Kartikay Khandelwal, Naman Goyal, Vishrav Chaudhary, Guillaume
  Wenzek, Francisco Guzm{\'a}n, Edouard Grave, Myle Ott, Luke Zettlemoyer, and
  Veselin Stoyanov. 2020.
\newblock Unsupervised cross-lingual representation learning at scale.
\newblock In \emph{Proceedings of the 58th Annual Meeting of the Association
  for Computational Linguistics}.

\bibitem[{De-Arteaga et~al.(2019)De-Arteaga, Romanov, Wallach, Chayes, Borgs,
  Chouldechova, Geyik, Kenthapadi, and Kalai}]{biosbias}
Maria De-Arteaga, Alexey Romanov, Hanna Wallach, Jennifer Chayes, Christian
  Borgs, Alexandra Chouldechova, Sahin Geyik, Krishnaram Kenthapadi, and
  Adam~Tauman Kalai. 2019.
\newblock Bias in bios: A case study of semantic representation bias in a
  high-stakes setting.
\newblock In \emph{Proceedings of the Conference on Fairness, Accountability,
  and Transparency}.

\bibitem[{Devlin et~al.(2019)Devlin, Chang, Lee, and Toutanova}]{bert}
Jacob Devlin, Ming-Wei Chang, Kenton Lee, and Kristina Toutanova. 2019.
\newblock \href {https://doi.org/10.18653/v1/N19-1423} {{BERT}: Pre-training of
  deep bidirectional transformers for language understanding}.
\newblock In \emph{Proceedings of the 2019 Conference of the North {A}merican
  Chapter of the Association for Computational Linguistics: Human Language
  Technologies, Volume 1 (Long and Short Papers)}, pages 4171--4186,
  Minneapolis, Minnesota. Association for Computational Linguistics.

\bibitem[{Dufter and Sch{\"u}tze(2020)}]{phillip_mbert}
Philipp Dufter and Hinrich Sch{\"u}tze. 2020.
\newblock \href {https://doi.org/10.18653/v1/2020.emnlp-main.358} {Identifying
  elements essential for {BERT}{'}s multilinguality}.
\newblock In \emph{Proceedings of the 2020 Conference on Empirical Methods in
  Natural Language Processing (EMNLP)}, pages 4423--4437, Online. Association
  for Computational Linguistics.

\bibitem[{Elazar et~al.(2021)Elazar, Ravfogel, Jacovi, and Goldberg}]{amnesic}
Yanai Elazar, Shauli Ravfogel, Alon Jacovi, and Yoav Goldberg. 2021.
\newblock Amnesic probing: Behavioral explanation with amnesic counterfactuals.
\newblock \emph{Transactions of the Association for Computational Linguistics},
  9:160--175.

\bibitem[{Gonen et~al.(2019)Gonen, Kementchedjhieva, and
  Goldberg}]{grammatical}
Hila Gonen, Yova Kementchedjhieva, and Yoav Goldberg. 2019.
\newblock How does grammatical gender affect noun representations in
  gender-marking languages?
\newblock In \emph{Proceedings of the 23rd Conference on Computational Natural
  Language Learning (CoNLL)}, Hong Kong, China.

\bibitem[{Gonen et~al.(2020)Gonen, Ravfogel, Elazar, and Goldberg}]{mbert}
Hila Gonen, Shauli Ravfogel, Yanai Elazar, and Yoav Goldberg. 2020.
\newblock It{'}s not {G}reek to m{BERT}: Inducing word-level translations from
  multilingual {BERT}.
\newblock In \emph{Proceedings of the Third BlackboxNLP Workshop on Analyzing
  and Interpreting Neural Networks for NLP}, Online.

\bibitem[{Karthikeyan et~al.(2020)Karthikeyan, Wang, Mayhew, and Roth}]{KWM20}
K~Karthikeyan, Zihan Wang, Stephen Mayhew, and Dan Roth. 2020.
\newblock Cross-lingual ability of multilingual bert: An empirical study.
\newblock In \emph{International Conference on Learning Representations}.

\bibitem[{Liang et~al.(2020)Liang, Dufter, and Sch{\"u}tze}]{philipp_debias}
Sheng Liang, Philipp Dufter, and Hinrich Sch{\"u}tze. 2020.
\newblock \href {https://doi.org/10.18653/v1/2020.coling-main.446} {Monolingual
  and multilingual reduction of gender bias in contextualized representations}.
\newblock In \emph{Proceedings of the 28th International Conference on
  Computational Linguistics}, pages 5082--5093, Barcelona, Spain (Online).
  International Committee on Computational Linguistics.

\bibitem[{Libovick{\'y} et~al.(2020)Libovick{\'y}, Rosa, and Fraser}]{LRF20}
Jind{\v{r}}ich Libovick{\'y}, Rudolf Rosa, and Alexander Fraser. 2020.
\newblock \href {https://doi.org/10.18653/v1/2020.findings-emnlp.150} {On the
  language neutrality of pre-trained multilingual representations}.
\newblock In \emph{Findings of the Association for Computational Linguistics:
  EMNLP 2020}, pages 1663--1674, Online. Association for Computational
  Linguistics.

\bibitem[{Liu et~al.(2020)Liu, Ott, Goyal, Du, Joshi, Chen, Levy, Lewis,
  Zettlemoyer, and Stoyanov}]{roberta}
Yinhan Liu, Myle Ott, Naman Goyal, Jingfei Du, Mandar Joshi, Danqi Chen, Omer
  Levy, Mike Lewis, Luke Zettlemoyer, and Veselin Stoyanov. 2020.
\newblock Roberta: A robustly optimized bert pretraining approach.
\newblock In \emph{ICLR}.

\bibitem[{May et~al.(2019)May, Wang, Bordia, Bowman, and
  Rudinger}]{gender-encoders}
Chandler May, Alex Wang, Shikha Bordia, Samuel~R. Bowman, and Rachel Rudinger.
  2019.
\newblock On measuring social biases in sentence encoders.
\newblock In \emph{Proceedings of the 2019 Conference of the North {A}merican
  Chapter of the Association for Computational Linguistics: Human Language
  Technologies, Volume 1 (Long and Short Papers)}, Minneapolis, Minnesota.
  Association for Computational Linguistics.

\bibitem[{Muller et~al.(2020)Muller, Sagot, and Seddah}]{MSS20}
Benjamin Muller, Benoˆıt Sagot, and Djame Seddah. 2020.
\newblock Can multilingual language models transfer to an unseen dialect? a
  case study on north african arabizi.
\newblock \emph{arXiv:2005.00318}.

\bibitem[{Peters et~al.(2018)Peters, Neumann, Iyyer, Gardner, Clark, Lee, and
  Zettlemoyer}]{elmo}
Matthew~E. Peters, Mark Neumann, Mohit Iyyer, Matt Gardner, Christopher Clark,
  Kenton Lee, and Luke Zettlemoyer. 2018.
\newblock \href {https://doi.org/10.18653/v1/N18-1202} {Deep contextualized
  word representations}.
\newblock In \emph{Proceedings of the 2018 Conference of the North {A}merican
  Chapter of the Association for Computational Linguistics: Human Language
  Technologies, Volume 1 (Long Papers)}, pages 2227--2237, New Orleans,
  Louisiana. Association for Computational Linguistics.

\bibitem[{Pires et~al.(2019)Pires, Schlinger, and Garrette}]{PSG19}
Telmo Pires, Eva Schlinger, and Dan Garrette. 2019.
\newblock \href {https://doi.org/10.18653/v1/P19-1493} {How multilingual is
  multilingual {BERT}?}
\newblock In \emph{Proceedings of the 57th Annual Meeting of the Association
  for Computational Linguistics}, pages 4996--5001, Florence, Italy.
  Association for Computational Linguistics.

\bibitem[{Ravfogel et~al.(2020)Ravfogel, Elazar, Gonen, Twiton, and
  Goldberg}]{inlp}
Shauli Ravfogel, Yanai Elazar, Hila Gonen, Michael Twiton, and Yoav Goldberg.
  2020.
\newblock \href {https://doi.org/10.18653/v1/2020.acl-main.647} {Null it out:
  Guarding protected attributes by iterative nullspace projection}.
\newblock In \emph{Proceedings of the 58th Annual Meeting of the Association
  for Computational Linguistics}, pages 7237--7256, Online. Association for
  Computational Linguistics.

\bibitem[{Ravfogel et~al.(2021)Ravfogel, Prasad, Linzen, and
  Goldberg}]{rc-counterfactuals}
Shauli Ravfogel, Grusha Prasad, Tal Linzen, and Yoav Goldberg. 2021.
\newblock \href {https://aclanthology.org/2021.conll-1.15} {Counterfactual
  interventions reveal the causal effect of relative clause representations on
  agreement prediction}.
\newblock In \emph{Proceedings of the 25th Conference on Computational Natural
  Language Learning}, pages 194--209, Online. Association for Computational
  Linguistics.

\bibitem[{Ravfogel et~al.(2022)Ravfogel, Twiton, Goldberg, and
  Cotterell}]{ravfogel2022linear}
Shauli Ravfogel, Michael Twiton, Yoav Goldberg, and Ryan Cotterell. 2022.
\newblock Linear adversarial concept erasure.
\newblock \emph{arXiv preprint arXiv:2201.12091}.

\bibitem[{Singh et~al.(2019)Singh, McCann, Socher, and Xiong}]{SMS19}
Jasdeep Singh, Bryan McCann, Richard Socher, and Caiming Xiong. 2019.
\newblock \href {https://doi.org/10.18653/v1/D19-6106} {{BERT} is not an
  interlingua and the bias of tokenization}.
\newblock In \emph{Proceedings of the 2nd Workshop on Deep Learning Approaches
  for Low-Resource NLP (DeepLo 2019)}, pages 47--55, Hong Kong, China.
  Association for Computational Linguistics.

\bibitem[{Wang et~al.(2019)Wang, Che, Guo, Liu, and Liu}]{WCG19}
Yuxuan Wang, Wanxiang Che, Jiang Guo, Yijia Liu, and Ting Liu. 2019.
\newblock \href {https://doi.org/10.18653/v1/D19-1575} {Cross-lingual {BERT}
  transformation for zero-shot dependency parsing}.
\newblock In \emph{Proceedings of the 2019 Conference on Empirical Methods in
  Natural Language Processing and the 9th International Joint Conference on
  Natural Language Processing (EMNLP-IJCNLP)}, pages 5721--5727, Hong Kong,
  China. Association for Computational Linguistics.

\bibitem[{Williams et~al.(2021)Williams, Cotterell, Wolf-Sonkin, Blasi, and
  Wallach}]{gg_ryan}
Adina Williams, Ryan Cotterell, Lawrence Wolf-Sonkin, Dami{\'a}n Blasi, and
  Hanna Wallach. 2021.
\newblock On the relationships between the grammatical genders of inanimate
  nouns and their co-occurring adjectives and verbs.
\newblock \emph{Transactions of the Association for Computational Linguistics},
  9:139--159.

\bibitem[{Wolf et~al.(2020)Wolf, Debut, Sanh, Chaumond, Delangue, Moi, Cistac,
  Rault, Louf, Funtowicz, Davison, Shleifer, von Platen, Ma, Jernite, Plu, Xu,
  Le~Scao, Gugger, Drame, Lhoest, and Rush}]{hugging}
Thomas Wolf, Lysandre Debut, Victor Sanh, Julien Chaumond, Clement Delangue,
  Anthony Moi, Pierric Cistac, Tim Rault, Remi Louf, Morgan Funtowicz, Joe
  Davison, Sam Shleifer, Patrick von Platen, Clara Ma, Yacine Jernite, Julien
  Plu, Canwen Xu, Teven Le~Scao, Sylvain Gugger, Mariama Drame, Quentin Lhoest,
  and Alexander Rush. 2020.
\newblock \href {https://doi.org/10.18653/v1/2020.emnlp-demos.6} {Transformers:
  State-of-the-art natural language processing}.
\newblock In \emph{Proceedings of the 2020 Conference on Empirical Methods in
  Natural Language Processing: System Demonstrations}, pages 38--45, Online.
  Association for Computational Linguistics.

\bibitem[{Wu and Dredze(2019)}]{WD19}
Shijie Wu and Mark Dredze. 2019.
\newblock Beto, bentz, becas: The surprising cross-lingual effectiveness of
  {BERT}.
\newblock In \emph{Proceedings of the 2019 Conference on Empirical Methods in
  Natural Language Processing and the 9th International Joint Conference on
  Natural Language Processing (EMNLP-IJCNLP)}, pages 833--844.

\bibitem[{Zhao et~al.(2020)Zhao, Mukherjee, Hosseini, Chang, and
  Hassan~Awadallah}]{mbios}
Jieyu Zhao, Subhabrata Mukherjee, Saghar Hosseini, Kai-Wei Chang, and Ahmed
  Hassan~Awadallah. 2020.
\newblock Gender bias in multilingual embeddings and cross-lingual transfer.
\newblock In \emph{Proceedings of the 58th Annual Meeting of the Association
  for Computational Linguistics}, Online. Association for Computational
  Linguistics.

\bibitem[{Zhao et~al.(2019)Zhao, Wang, Yatskar, Cotterell, Ordonez, and
  Chang}]{gender-ctx}
Jieyu Zhao, Tianlu Wang, Mark Yatskar, Ryan Cotterell, Vicente Ordonez, and
  Kai-Wei Chang. 2019.
\newblock Gender bias in contextualized word embeddings.
\newblock In \emph{Proceedings of the 2019 Conference of the North {A}merican
  Chapter of the Association for Computational Linguistics: Human Language
  Technologies, Volume 1 (Long and Short Papers)}, Minneapolis, Minnesota.

\bibitem[{Zhou et~al.(2019)Zhou, Shi, Zhao, Huang, Chen, Cotterell, and
  Chang}]{ZSZ19}
Pei Zhou, Weijia Shi, Jieyu Zhao, Kuan-Hao Huang, Muhao Chen, Ryan Cotterell,
  and Kai-Wei Chang. 2019.
\newblock \href {https://doi.org/10.18653/v1/D19-1531} {Examining gender bias
  in languages with grammatical gender}.
\newblock In \emph{Proceedings of the 2019 Conference on Empirical Methods in
  Natural Language Processing and the 9th International Joint Conference on
  Natural Language Processing (EMNLP-IJCNLP)}, pages 5276--5284, Hong Kong,
  China. Association for Computational Linguistics.

\bibitem[{Zmigrod et~al.(2019)Zmigrod, Mielke, Wallach, and
  Cotterell}]{spanishbias}
Ran Zmigrod, Sabrina~J. Mielke, Hanna Wallach, and Ryan Cotterell. 2019.
\newblock \href {https://aclanthology.org/P19-1161} {Counterfactual data
  augmentation for mitigating gender stereotypes in languages with rich
  morphology}.
\newblock In \emph{Proceedings of the 57th Annual Meeting of the Association
  for Computational Linguistics}, Florence, Italy. Association for
  Computational Linguistics.

\end{thebibliography}
\bibliographystyle{acl_natbib}

\end{document}